\documentclass[USenglish]{article}	

\usepackage[utf8]{inputenc}				
\usepackage[big,online]{dgruyter}	
\usepackage{lmodern} 
\usepackage{microtype}
\usepackage[numbers,square,sort&compress]{natbib}

\usepackage{amssymb}
\usepackage{amsmath}
\usepackage{hyperref}

\usepackage{authblk}

\usepackage{algorithm}
\usepackage{algorithmic}
\makeatletter
\providecommand{\plist@algorithm}{Algorithm\space}
\makeatother

\theoremstyle{dgthm}

\theoremstyle{dgdef}

\begin{document}

\makeatletter
\sbox\dg@wordmark{}
\makeatother
	
	\articletype{Research Article}

\title{Geological Field Restoration \\through the Lens of Image Inpainting}
\runningtitle{Geological Field Restoration through the Lens of Image Inpaintin}

\author*[1]{Vladislav~Trifonov}
\author[2]{Ivan~Oseledets}
\author[3]{Ekaterina~Muravleva}
\runningauthor{V.~Trifonov et al.}
\affil[1]{\protect\raggedright 
Skoltech, AIC and Sberbank of Russia, AI4S Center, Moscow, Russia, e-mail: vladtrifono@gmail.com}
\affil[2]{\protect\raggedright 
AIRI and Skoltech, AIC, Moscow, Russia, e-mail: oseledets@airi.net}
\affil[3]{\protect\raggedright 
Sberbank of Russia, AI4S Center and Skoltech, AIC, Moscow, Russia, e-mail: e.muravleva@skoltech.ru}


	
\abstract{We study an ill-posed problem of geological field reconstruction under limited observations. Engineers often have to deal with the problem of reconstructing the subsurface geological field from sparse measurements such as exploration well data. Inspired by image inpainting, we model this partially observed spatial field as a multidimensional tensor and recover missing values by enforcing a global low rank structure together with spatial smoothness. We solve the resulting optimization via the Alternating Direction Method of Multipliers. On the SPE10 model 2 benchmark, this deterministic approach yields consistently lower relative squared error than ordinary kriging across various sampling densities and produces visually coherent reconstructions.}

\keywords{tensor completion, geostatistics, image inpainting, spatial interpolation}

\maketitle

\section{Introduction}



Reconstructing a subsurface property field from sparse well measurements is a central task in geoscience with applications in resource exploration, groundwater modeling, environmental monitoring, and resource extraction. Because only narrow well trajectories are observed, interpolating the full 3D volume is ill-posed: many possible fields fit the limited data. A common approach is kriging (Gaussian process regression, GPR), which interpolates based on a specified covariance (variogram) model and yields best linear unbiased predictions under that model. However, estimating reliable variograms from very sparse data is difficult and often requires substantial expert tuning. Moreover, in practice well data are often far more sparse than the assumed spatial correlation range required for reliable GPR interpolation.


In contrast, we frame geological field reconstruction as an image inpainting problem. A single 3D property field is a three-way array $\mathcal{X}\in\mathbb{R}^{I\times J\times K}$, partially observed along well trajectories. The wellbore data along the $z$-axis can be regarded as a two-dimensional image with additional channel dimensions, analogous to color channels in image processing. This perspective leverages recent machine-learning advances in image reconstruction. We adopt a classical approach from multilinear algebra and solve the problem via tensor completion. The task is formulated as an optimization problem of finding a low rank representation of the data tensor, which we solve using the Alternating Direction Method of Multipliers (ADMM).

Our contributions are as follows: 
\begin{itemize}
    \item We propose a new perspective on geological field reconstruction by formulating it as an image inpainting task.
    \item We adopt tensor completion to reconstruct geological data tensors.
    \item We demonstrate that the proposed tensor completion approach achieves superior reconstruction performance on a complex 3D geological formation compared with ordinary kriging, while requiring no explicit prior domain knowledge.
\end{itemize}

\medskip\noindent\textbf{Related work~~}Low-rank completion has been effectively applied to seismic data interpolation~\citep{andersen2004structure, acar2005modeling, acar2006collective, ely20135d} and petrophysical property estimation~\citep{insuasty2017low}. The most similar application to geological field restoration was presented in~\citep{syed2022low}. Our work differs in several ways:
\begin{itemize}
    \item We formulate explicit optimization objectives and derive specific algorithms, rather than treating tensor completion as a black-box. This formalism also allows incorporating additional domain- or method-specific knowledge.
    \item In contrast to~\citep{syed2022low}, which removed small artificial gaps from complete fields, we address realistic scenarios with truly sparse observations (i.e., well data).
    \item We explicitly benchmark our reconstruction performance against the well-established baseline of ordinary kriging. While~\citep{syed2022low} uses a fluid-flow recovery simulation as the metric, implementing such a simulation with very sparse data requires substantially more engineering effort. We argue that the primary evaluation should be reconstruction accuracy against a baseline interpolator, and we leave fluid recovery simulation as a metric for future work.
\end{itemize}

\section{Gaussian process regression baseline}

In geostatistics, kriging is the standard method for spatial interpolation of volumetric data from observed data points. Kriging treats the quantity of interest $Z(u)$ at location $u$ as a realization of a random field and constructs a linear predictor for unsampled locations. The ordinary kriging~\citep{cressie1990origins} estimator at a new point $u_0$ is given by
\begin{equation*}
    \hat{Z}(u_0) = \sum_{i}^{n} \lambda_i Z(u_i)~,
\end{equation*}
where the weights $\lambda_i$ minimize the prediction variance under a covariance model estimated from the observed data. Kriging is widely used across geoscience \citep{delhomme1978kriging}. In addition to engineering applications, GPR has many applications in machine learning~\citep{seeger2004gaussian}.

The Stanford Geostatistical Modeling Software (SGeMS) is used for baseline generation~\citep{remy2009applied}. SGeMS is open-source software that implements all of the most common tools for spatial statistics. Kriging is a nonparametric technique that uses observed data to estimate the mean and covariance parameters of the distribution, and then uses this estimated distribution to make predictions for new inputs \citep{cressie1990origins}. The linear system for simple kriging built from a covariance matrix is~\citep{remy2009applied}:
\begin{equation*}
    \begin{bmatrix}
        C_{11} & \cdots & C_{1n} \\
        \vdots & \ddots & \vdots \\
        C_{n1} & \cdots & C_{nn}
    \end{bmatrix}
    \begin{bmatrix}
        \lambda_1 \\ \vdots \\ \lambda_n
    \end{bmatrix}
    =
    \begin{bmatrix}
        C_{1u} \\ \vdots \\ C_{nu}
    \end{bmatrix}~,
\end{equation*}
where $C_{ij}$ denotes the covariance between data points $i$ and $j$, and $ C_{iu}$ denotes the covariance between data point $i$ and the unsampled location $u$. Both are derived from the assumed stationary covariance model:
\begin{equation*}
    C_{ij} = \texttt{Cov}\big(Z(u_i), Z(u_j)\big) = C(0) - \gamma(u_i - u_j)~,
\end{equation*}
where $C(0) = \texttt{Var}\big(Z(u)\big)$ is a stationary variance, $\gamma(h)$ is a corresponding stationary half-variogram model:
\begin{equation*}
    2\gamma(h) = \texttt{Var}\big( Z(u) - Z(u+h) \big)~.
\end{equation*}

Several kriging variants are used in geostatistics, including ordinary kriging (which assumes an unknown constant local mean), simple kriging (known constant mean), and co-kriging (jointly interpolating multiple correlated properties). In this work, we use ordinary kriging as a baseline interpolator.

\section{Image inpainting for geological fields}

Image inpainting is an image processing task that involves filling in missing or damaged regions of an image using the surrounding information. This task comes from fields such as computer vision, machine learning, and signal processing~\citep{elharrouss2020image}. The reconstructed missing parts should be visually coherent with the rest of the image.

Tensors are multidimensional arrays that generalize scalars, vectors, and matrices to higher dimensions. A real-valued tensor of order $N$ is denoted $\mathcal{A} \in \mathbb{R}^{I_0 \times I_1 \times \dots \times I_{N-1}}$, where $I_n$ is the size of the $n$-th dimension.

Treating spatial field data as tensors enables alternative modeling approaches. Tensor completion methods typically use the low rank representation assumption to efficiently estimate missing entries. Unlike matrices, tensors do not have a single rank definition. The notion of rank depends on the chosen tensor decomposition, leading to different modeling strategies. 

A common convex relaxation of the Tucker rank is to minimize the sum of nuclear norms of the unfoldings~\citep{cai2010singular}: $\min_{\mathcal{X}} \sum_{n=0}^{N-1} \Vert X_{(n)} \Vert_*$, where $\Vert X_{(n)} \Vert_*$ is the nuclear norm (sum of singular values) of the mode-$n$ unfolding. For $\mathcal{X} \in \mathbb{R}^{I_0 \times I_1 \times \dots \times I_{N-1}}$, the mode-$n$ unfolding ${X_{(n)}} \in \mathbb{R}^{I_n \times (I_0\,\cdot\dots\,\cdot I_{n - 1}\,\cdot I_{n + 1}\,\cdot \dots \,\cdot I_{N-1})}$ has first dimension $I_n$ and second dimension equal to the product of the remaining sizes. 

\medskip\noindent\textbf{Low rank tensor completion~~}We aim to reconstruct the estimate $\hat{\mathcal{X}}$ from a partially observed tensor $\hat{\mathcal{Y}}$ of the true tensor $\mathcal{X}^*$, by solving:
\begin{equation}
\label{opt_problem}
    \arg\min_\mathcal{X} \Vert \mathcal{P}_{\Omega}(\mathcal{X}) - \mathcal{P}_{\Omega}(\mathcal{Y}) \Vert^2_F~,
\end{equation}
where $\Omega$ is a binary tensor with the same dimensions as $\mathcal{Y}$ indicating observed elements, and $\Vert \cdot \Vert_F$ is the Frobenius norm. The projection operator $\mathcal{P}_{\Omega}$ returns the tensor element if it is present and zero otherwise:
\begin{equation*}
    \mathcal{P}_{\Omega}(\mathcal{Y}) =
    \begin{cases}
    y_{i_0, \dots,  i_{N-1}}, & \text{for}~(i_0, \dots,  i_{N-1}) \in \Omega \\
    0~, & \text{otherwise}
    \end{cases}~.
\end{equation*}
Note that $\mathcal{P}_{\Omega}^{\perp}$ denotes the complementary projection onto unobserved entries.

Problem~\eqref{opt_problem} is ill-posed and requires additional constraints. One must either estimate the tensor rank or include a rank-minimization surrogate in the objective~\eqref{opt_problem}. We adopt the following convex relaxation:
\begin{equation}
    \label{nuclear_norm_sum}
    \min_{\mathcal{X}} \sum_{n=0}^{N-1} \Vert X_{(n)} \Vert_* \quad \text{s.t.} \quad \mathcal{P}_{\Omega}(\mathcal{X}) = \mathcal{P}_{\Omega}(\mathcal{Y})~.
    \end{equation}

Intuitively, this objective encourages $\mathcal{X}$ to have low rank across all unfoldings, while exactly matching the observed data. The low-rank prior reflects the assumption that the underlying field has limited intrinsic degrees of freedom, consistent with the structured nature of geological formations. We solve this problem using the ADMM algorithm.

We introduce an auxiliary variable $\mathcal{Z}$ and solve
\begin{equation} \label{admm_obj}
    \min_{\mathcal{X}, \mathcal{Z}} \Vert \mathcal{Z}\Vert_* \quad \text{s.t.} \quad \mathcal{X} = \mathcal{Z}, \quad \mathcal{P}_\Omega(\mathcal{X}) = \mathcal{P}_\Omega(\mathcal{Y})~.
\end{equation}

The ADMM iterations alternate between (i) singular value soft-thresholding (SVT) on each unfolding, promoting low rank, and (ii) data-consistency projection that restores observed entries using $\mathcal{P}_\Omega$. We denote by $\text{SVT}(A, \beta)$ the singular value soft-thresholding of matrix $A$ with parameter $\beta$. If $A = U\Sigma V^\top$ is the SVD of $A$, then $\text{SVT}(A, \beta) = U S V^\top$, where $S$ has diagonal entries $S_{ii} = \max{\{\sigma_i - \beta, 0\}}$. We use standard dual updates with penalty parameter $\rho$.

\medskip\noindent\textbf{Spatial smoothness~~}In our experiments, the basic ADMM approach converged slowly and often failed to recover a satisfactory reconstruction. To overcome this, we employed the following variant based on extensive literature~\citep{liu2012tensor, wang2016tensor, chen2020nonconvex}.

First, we treat each mode's unfolding as an independent subproblem, then average the three mode-wise solutions at each iteration to obtain the updated tensor. This enforces consistency across the low-rank representations of the different modes. Second, we incorporate a graph-Laplacian regularization on the horizontal dimensions (modes $0$ and $1$), to encourage the lateral continuity typical of layered media in the $x$-$y$ plane. Thus, instead of objective~\eqref{nuclear_norm_sum}, we solve the regularized problem:
\begin{equation*}
    \min_{\mathcal{X}} \sum_{n=0}^{2} \Vert X_{(n)} \Vert_* + \frac{\beta}{2} \sum_{n=0}^{1} \Vert D_n X_{(n)} \Vert_F^2 \quad \text{s.t.} \quad \mathcal{P}_\Omega(\mathcal{X}) = \mathcal{P}_\Omega(\mathcal{Y})~.
\end{equation*}
Here $D_0$ and $D_1$ are discrete difference operators along horizontal axes and $\beta$ controls smoothing strength. This adds a mild prior that adjacent lateral cells should not vary abruptly unless required by the data.

See Algorithm~\ref{smoothed_admm_alg} for pseudo-code. The functions $\text{unfold}_n$ and $\text{fold}_n$ denote the unfolding and folding operators, respectively. For example, if $\mathcal{A} \in \mathbb{R}^{I_0 \times I_1 \times I_2 \times I_3}$, then $\text{unfold}_2(\mathcal{A}) \in \mathbb{R}^{I_2 \times (I_0 \cdot I_1 \cdot I_3)}$ rearranges the tensor into a matrix along mode 2. By contrast, a slice such as $\mathcal{A}{[...,3]} \in \mathbb{R}^{I_0 \times I_1 \times I_2}$ refers to fixing the last index at 3.


\begin{algorithm}[!h]
\caption{Graph Laplacian Smoothed Nuclear Norm Minimization for 3D Tensor Completion}
\begin{algorithmic}[1]
\label{smoothed_admm_alg}
\REQUIRE Observed tensor $\mathcal{Y} \in \mathbb{R}^{I \times J \times K}$, parameters $\alpha, \beta, \rho$, maximum iterations $K$
\ENSURE Reconstructed tensor $\hat{\mathcal{X}}$

\STATE \textbf{Initialize:} $\mathcal{X}^0 = \text{stack}(\mathcal{Y}, \mathcal{Y}, \mathcal{Y}),~\mathcal{Z}^0 = 0,~\mathcal{T}^0 = 0, \quad \mathcal{X}^0, \mathcal{Z}^0, \mathcal{T}^0\in \mathbb{R}^{I \times J \times K \times 3}$
\STATE \textbf{Precompute:}
$\mathbf{inv}_0 = (\beta \mathbf{D}_0^\top \mathbf{D}_0 + \rho \mathbf{I})^{-1}, \quad
\mathbf{inv}_1 = (\beta \mathbf{D}_1^\top \mathbf{D}_1 + \rho \mathbf{I})^{-1}$

\FOR{$k = 0$ to $K-1$}
    \STATE \textit{// Mode-$n$ singular value thresholding:}
    \FOR{$n = 0$ to $2$}
        \STATE $\mathbf{W}_{(n)} = \text{unfold}_n \left( \mathcal{X}^{k}_{[..., n]} + \frac{1}{\rho} \mathcal{T}^{k}_{[..., n]} \right)$
        \STATE $\mathbf{Z}_{(n)}^{k+1} = \text{SVT}(\mathbf{W}_{(n)}, \frac{\alpha}{\rho})$
        \STATE $\mathcal{Z}^{k+1}_{[..., n]} = \text{fold}_n \left( \mathbf{Z}_{(n)}^{k+1} \right)$
    \ENDFOR

    \STATE \textit{// Graph-smooth updates with intermediate tensor $\mathcal{V}\in \mathbb{R}^{I \times J \times K \times 3}$:}
    \STATE $\mathbf{rhs}_0 = \text{unfold}_0 \left( \rho \mathcal{Z}^{k+1}_{[..., 0]} - \mathcal{T}^{k}_{[..., 0]} \right)$
    \STATE $\mathbf{rhs}_1 = \text{unfold}_1 \left( \rho \mathcal{Z}^{k+1}_{[..., 1]} - \mathcal{T}^{k}_{[..., 1]} \right)$

    \STATE $\mathcal{V}_{[..., 0]} = \text{fold}_0 \left( \mathbf{inv}_0 \cdot \mathbf{rhs}_0 \right)$
    \STATE $\mathcal{V}_{[..., 1]} = \text{fold}_1 \left( \mathbf{inv}_1 \cdot \mathbf{rhs}_1 \right)$
    \STATE $\mathcal{V}_{[..., 2]} = \mathcal{Z}^{k+1}_{[..., 2]} - \frac{1}{\rho} \mathcal{T}^{k}_{[..., 2]}$

    \STATE \textit{// Update $\mathcal{X}^{k+1}$ ensuring consistentcy with the known data:}
    \FOR{$n=0$ to $2$}
        \STATE $\mathcal{X}^{k+1}_{[..., n]} = \mathcal{P}_{\Omega}^{\perp}(\mathcal{V}_{[..., n]}) + \mathcal{P}_{\Omega}(\mathcal{Y})$
    \ENDFOR

    \STATE \textit{// Dual variable update:}
    \STATE $\mathcal{T}^{k+1} = \mathcal{T}^k + \rho (\mathcal{X}^{k+1} - \mathcal{Z}^{k+1})$

    \STATE \textit{// Compute intermediate reconstruction:}
    \STATE $\hat{\mathcal{X}}^{k+1} = \frac{1}{3} \sum_{n=0}^2 \mathcal{X}^{k+1}_{[..., n]}$
    \STATE \textit{// Stack reconstruction for next iteration:}
    \STATE $\mathcal{X}^{k+1} = \text{stack}(\hat{\mathcal{X}}^{k+1}, \hat{\mathcal{X}}^{k+1}, \hat{\mathcal{X}}^{k+1})$
\ENDFOR

\RETURN $\mathcal{P}_{\Omega}^{\perp}(\hat{\mathcal{X}}) + \mathcal{P}_{\Omega}(\mathcal{Y})$

\end{algorithmic}
\end{algorithm}

\section{Computational experiments}



\medskip\noindent\textbf{Overestimated results with kriging~~}The first step in interpolating the field with kriging is variogram estimation. This is a difficult task due to sparse and noisy data. Moreover, kriging requires the known data points to lie within a given correlation radius, which is difficult to achieve in practice since wells can be kilometers apart.

In our experiments, we gave kriging the advantage of the true variogram (i.e., one computed from the entire field). This dramatically improves kriging's performance — the very sparse well data alone would not permit fitting a variogram model as well as the true one. Furthermore, kriging performance can also be limited by noise and non-stationary structures.

Finally, synthetic datasets, like the one we used in this paper, are created as Gaussian random fields (GRFs). In such cases, Gaussian process regression can employ a covariance function that closely matches the generating model.

Even with these favorable conditions for kriging, our tensor completion approach still achieves higher accuracy and better visual coherence, all without using any prior knowledge of the field.

\begin{figure*}[!h]
\normalsize
\centering
    \caption{3D porosity field of SPE10 model 2 geological benchmark.}
    \begin{center}
        \includegraphics[width=.5\textwidth]{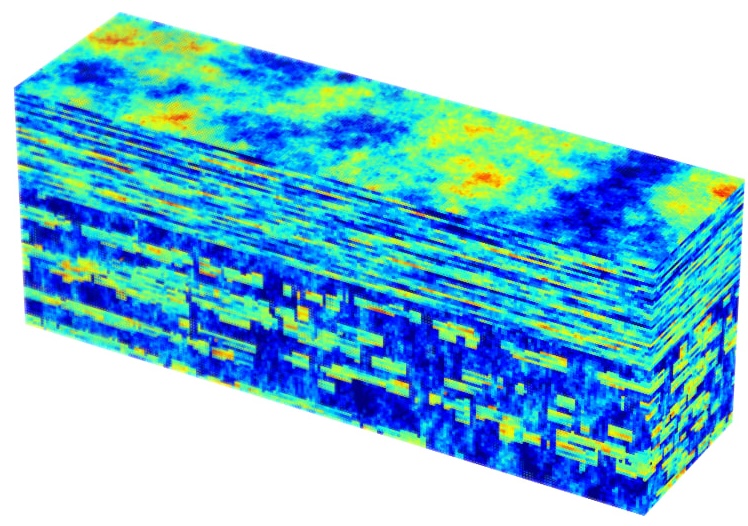}
    \end{center}
    \label{fig:spe10_model2}
    \vspace*{4pt}
\end{figure*}

\begin{figure*}[!h]
\normalsize
\centering
    \caption{Example of well data available during reconstruction process, $100$ wells.}
    \begin{center}
        \includegraphics[width=.6\textwidth]{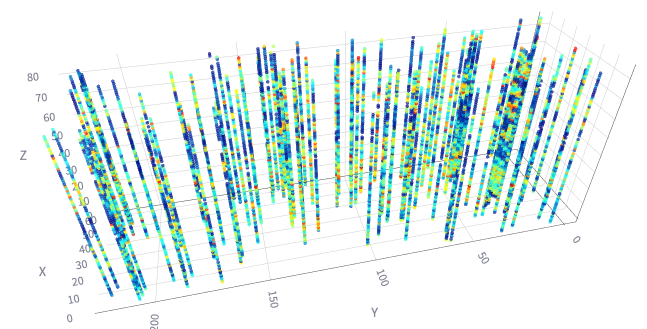}
    \end{center}
    \label{fig:well_samples}
    \vspace*{4pt}
\end{figure*}

\medskip\noindent\textbf{Geological field benchmark~~}The performance of the proposed approach is demonstrated on a porosity field from the well-known benchmark SPE10 model 2 \citep{christie2001tenth}. Figure~\ref{fig:spe10_model2} shows the full 3D porosity field of SPE10 model 2, and Figure~\ref{fig:well_samples} illustrates an example of the available well data (well locations and values) used for reconstruction. The SPE10 model 2 field was generated as a 3D GRF with about $1.12$ million cells.

Relevant characteristics of the SPE10 model 2 are:
\begin{itemize}
    \item Two different geological formations: (i) the upper part of the model represents a prograding nearshore environment; (ii) the lower part is fluvial, with clearly visible channels.
    \item Model size: $[1200 \times 2200 \times 170]$ ft.
    \item Cartesian grid size: $[60 \times 220 \times 85]$ cells.
    \item Cell size: $[20 \times 20 \times 2]$ ft.
\end{itemize}

\medskip\noindent\textbf{Metric~~}Recovery performance is measured by relative squared error (RSE) on only the unobserved (predicted) cells:
\begin{equation*}
     \text{RSE} = \frac{\Vert\mathcal{P}_{\Omega}^{\perp}(\mathcal{\hat{X}}) - \mathcal{P}_{\Omega}^{\perp}(\mathcal{Y})\Vert_F}{\Vert\mathcal{P}_{\Omega}^{\perp}(\mathcal{Y})\Vert_F}~.
\end{equation*}

\medskip\noindent\textbf{Reconstruction results~~}We reconstruct the field with different percentages of available data (i.e., number of wells) using tensor completion and compare the reconstructions to those obtained by kriging. For each well count, we randomly sample well positions simulating different exploration well configurations. The RSE results are summarized in Table~\ref{tab:results}. As shown in the table, tensor completion consistently achieves lower RSE than kriging for all cases. We report the mean RSE and standard deviation over 50 independent runs, each corresponding to a different random selection of well locations. Details on the experimental parameters are provided in Appendix~\ref{app:exp_details}.

\begin{figure*}[!h]
    \normalsize
    \centering
    \caption{Reconstruction results of the porosity field from SPE10 model 2. Cross-section along the $z$-axis at $z = 12$ with $500$ wells. Left to right: (i) ground truth from SPE10 model 2; (ii) kriging; (iii) tensor completion; (iv) well mask.}
        \begin{center}
            \includegraphics[width=.8\textwidth]{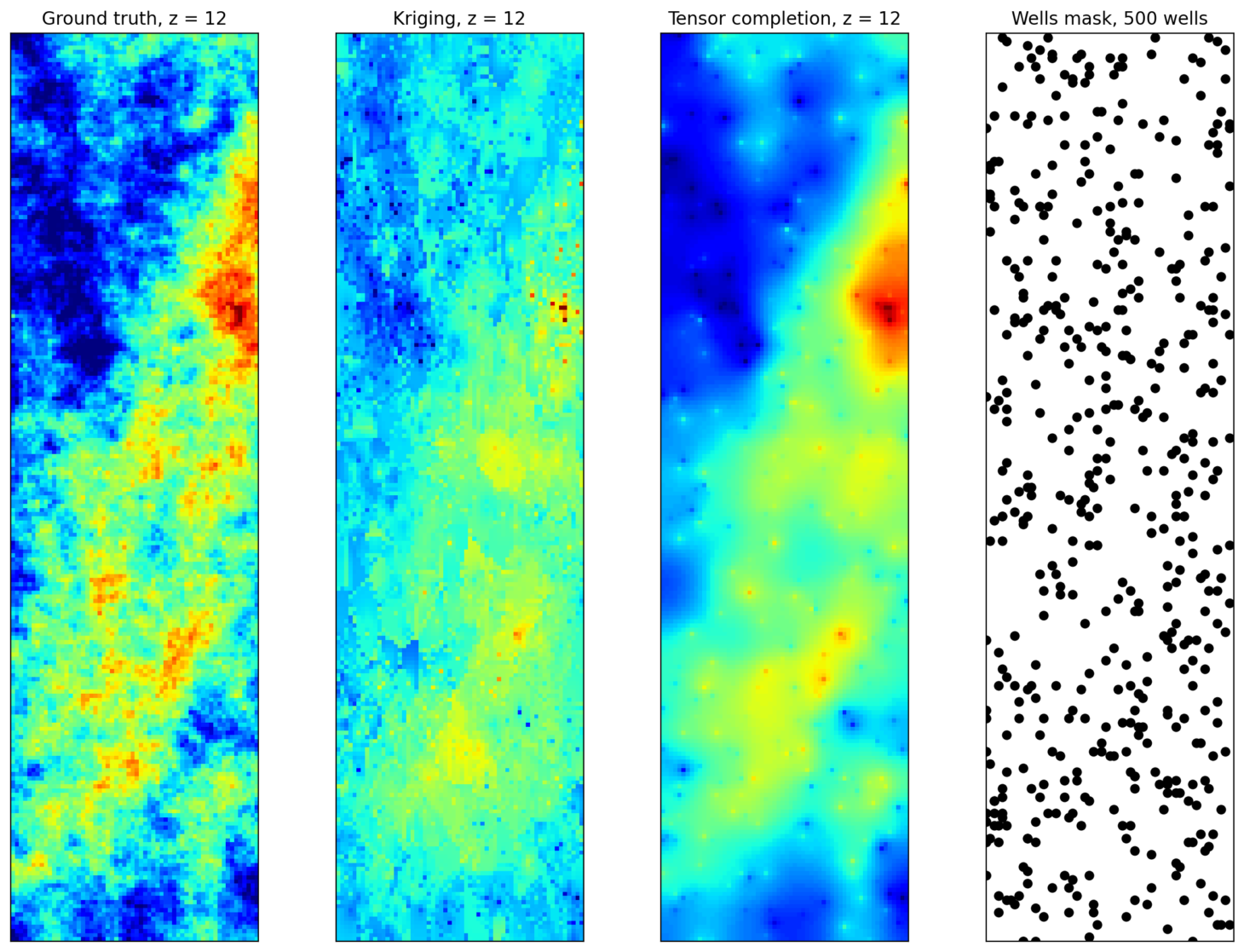}
        \end{center}
        \label{fig:rec_wells500_z12}
        \vspace*{4pt}
\end{figure*}

Beyond the numerical improvements, our low-rank approach also offers a qualitative advantage: it tends to preserve layer continuity (Figure~\ref{fig:rec_wells500_z12}). These trends are consistent for both the upper and lower formations. For example, kriging tends to over-smooth the upper formation (Appendix~\ref{app:rec_z12},~\ref{app:rec_z27}), while tensor completion is able to distinguish the major geological regions. Our method also better captures the fluvial channel patterns (Appendix~\ref{app:rec_z50},~\ref{app:rec_z75}), filling in riverbed details more coherently and producing smoother boundaries between regions. This coherence can benefit subsequent flow simulations, as rapid cell-to-cell property changes (e.g., in permeability) often degrade the stability of numerical solvers. With very sparse data, kriging can produce noisy, piecewise-constant artifacts in the interpolation, while the low-rank completion yields smoother fields that still honor the large-scale patterns.

\begin{table*}[!h]
    \caption{Mean and standard deviation of the RSE for reconstruction results with different numbers of wells, averaged over $50$ runs. Lower is better. Bold is the best.}
    \centering
    \begin{tabular}{cccc}
        Number of wells & Active cells, \% & RSE for kriging & RSE for tensor completion \\
        \midrule
        $100$ & $0.8$ & $0.476{\scriptstyle \pm 0.0020}$ & $\textbf{0.406}{\scriptstyle \pm 0.0066}$ \\
        $300$ & $2.3$ & $0.436{\scriptstyle \pm 0.0017}$ & $\textbf{0.351}{\scriptstyle \pm 0.0028}$ \\
        $500$ & $3.8$ & $0.416{\scriptstyle \pm 0.0012}$ & $\textbf{0.330}{\scriptstyle \pm 0.0022}$ \\
        $700$ & $5.3$ & $0.402{\scriptstyle \pm 0.0012}$ & $\textbf{0.319}{\scriptstyle \pm 0.0028}$ \\
    \end{tabular}
    \label{tab:results}
\end{table*}






\section {Conclusion and prospects}

The tensor completion algorithm offers an effective approach to recovering spatial fields. The proposed nuclear-norm minimization approach achieves higher reconstruction quality than ordinary kriging with SGeMS. We believe that the problem of reconstructing geological fields is closely related to matrix completion and image inpainting. By framing the problem in this manner, we can leverage robust convex optimization techniques. Our experiments demonstrate the method’s applicability in a geoscience context.


By enforcing a low-rank structure, we assume that the 3D geological field varies smoothly or exhibits similar patterns along different axes. This assumption is often reasonable for stratified subsurface media, where even a small number of wells can suffice to infer the full field when GPR would likely fail. However, if the geology is highly irregular or contains high-frequency variation (e.g., complex fracture networks), the low-rank approximation may break down. In such cases, kriging may capture small-scale details better, provided those features are directly sampled by wells. This makes the reliable recovery of fine-scale structures an open challenge for both kriging and tensor completion.

The main drawback of our method is the lack of any native uncertainty quantification. Kriging, by contrast, explicitly provides prediction uncertainty, which is critical for decision-making (e.g., in assessing drilling risk). Our formulation currently yields only a single best estimate. A natural extension would be to adopt a Bayesian tensor completion approach~\cite{shi2023bayesian} to quantify uncertainty.

When multiple attributes (e.g., porosity and permeability) are available, one can use co-kriging to interpolate them simultaneously by modeling their joint spatial covariance since such properties are typically correlated. Likewise, the tensor completion method can handle multiple properties by extending to a higher-order joint tensor, with an additional dimension introduced for each property. We leave the exploration of this extension for future work.

\bibliographystyle{abbrv}
\bibliography{tenscomp_jiip}

@article{cressie1990origins,
  title={The origins of kriging},
  author={Cressie, Noel},
  journal={Mathematical geology},
  volume={22},
  pages={239--252},
  year={1990},
  publisher={Springer}
}

@article{seeger2004gaussian,
  title={Gaussian processes for machine learning},
  author={Seeger, Matthias},
  journal={International journal of neural systems},
  volume={14},
  number={02},
  pages={69--106},
  year={2004},
  publisher={World Scientific}
}

@article{delhomme1978kriging,
  title={Kriging in the hydrosciences},
  author={Delhomme, Jean Pierre},
  journal={Advances in water resources},
  volume={1},
  number={5},
  pages={251--266},
  year={1978}
}

@book{remy2009applied,
  title={Applied geostatistics with SGeMS: A user's guide},
  author={Remy, Nicolas and Boucher, Alexandre and Wu, Jianbing},
  year={2009},
  publisher={Cambridge University Press}
}

@article{christie2001tenth,
  title={Tenth SPE comparative solution project: A comparison of upscaling techniques},
  author={Christie, Michael Andrew and Blunt, Martin J},
  journal={SPE Reservoir Evaluation \& Engineering},
  volume={4},
  number={04},
  pages={308--317},
  year={2001},
  publisher={SPE}
}

@software{jax2018github,
  author = {James Bradbury and Roy Frostig and Peter Hawkins and Matthew James Johnson and Chris Leary and Dougal Maclaurin and George Necula and Adam Paszke and Jake Vander{P}las and Skye Wanderman-{M}ilne and Qiao Zhang},
  title = {{JAX}: composable transformations of {P}ython+{N}um{P}y programs},
  url = {http://github.com/jax-ml/jax},
  version = {0.3.13},
  year = {2018},
}

@article{shi2023bayesian,
  title={Bayesian methods in tensor analysis},
  author={Shi, Yiyao and Shen, Weining},
  journal={arXiv preprint arXiv:2302.05978},
  year={2023}
}

@article{liu2012tensor,
  title={Tensor completion for estimating missing values in visual data},
  author={Liu, Ji and Musialski, Przemyslaw and Wonka, Peter and Ye, Jieping},
  journal={IEEE transactions on pattern analysis and machine intelligence},
  volume={35},
  number={1},
  pages={208--220},
  year={2012},
  publisher={IEEE}
}

@article{wang2016tensor,
  title={Tensor completion by alternating minimization under the tensor train (TT) model},
  author={Wang, Wenqi and Aggarwal, Vaneet and Aeron, Shuchin},
  journal={arXiv preprint arXiv:1609.05587},
  year={2016}
}

@article{syed2022low,
  title={Low-rank tensors applications for dimensionality reduction of complex hydrocarbon reservoirs},
  author={Syed, Fahad Iqbal and Muther, Temoor and Dahaghi, Amirmasoud Kalantari and Negahban, Shahin},
  journal={Energy},
  volume={244},
  pages={122680},
  year={2022},
  publisher={Elsevier}
}

@inproceedings{insuasty2017low,
  title={Low-dimensional tensor representations for the estimation of petrophysical reservoir parameters},
  author={Insuasty, E and Van den Hof, PM and Weiland, S and Jansen, Jan Dirk},
  booktitle={SPE Reservoir Simulation Conference},
  pages={D021S005R006},
  year={2017},
  organization={SPE}
}

@article{andersen2004structure,
  title={Structure-seeking multilinear methods for the analysis of fMRI data},
  author={Andersen, Anders H and Rayens, William S},
  journal={NeuroImage},
  volume={22},
  number={2},
  pages={728--739},
  year={2004},
  publisher={Elsevier}
}

@inproceedings{acar2005modeling,
  title={Modeling and multiway analysis of chatroom tensors},
  author={Acar, Evrim and Camtepe, Seyit A and Krishnamoorthy, Mukkai S and Yener, B{\"u}lent},
  booktitle={Intelligence and Security Informatics: IEEE International Conference on Intelligence and Security Informatics, ISI 2005, Atlanta, GA, USA, May 19-20, 2005. Proceedings 3},
  pages={256--268},
  year={2005},
  organization={Springer}
}

@inproceedings{acar2006collective,
  title={Collective sampling and analysis of high order tensors for chatroom communications},
  author={Acar, Evrim and Camtepe, Seyit A and Yener, B{\"u}lent},
  booktitle={International Conference on Intelligence and Security Informatics},
  pages={213--224},
  year={2006},
  organization={Springer}
}

@incollection{ely20135d,
  title={5D and 4D pre-stack seismic data completion using tensor nuclear norm (TNN)},
  author={Ely, Gregory and Aeron, Shuchin and Hao, Ning and Kilmer, Misha E},
  booktitle={SEG Technical Program Expanded Abstracts 2013},
  pages={3639--3644},
  year={2013},
  publisher={Society of Exploration Geophysicists}
}

@article{elharrouss2020image,
  title={Image inpainting: A review},
  author={Elharrouss, Omar and Almaadeed, Noor and Al-Maadeed, Somaya and Akbari, Younes},
  journal={Neural Processing Letters},
  volume={51},
  pages={2007--2028},
  year={2020},
  publisher={Springer}
}

@article{cai2010singular,
  title={A singular value thresholding algorithm for matrix completion},
  author={Cai, Jian-Feng and Cand{\`e}s, Emmanuel J and Shen, Zuowei},
  journal={SIAM Journal on optimization},
  volume={20},
  number={4},
  pages={1956--1982},
  year={2010},
  publisher={SIAM}
}

@article{chen2020nonconvex,
  title={A nonconvex low-rank tensor completion model for spatiotemporal traffic data imputation},
  author={Chen, Xinyu and Yang, Jinming and Sun, Lijun},
  journal={Transportation Research Part C: Emerging Technologies},
  volume={117},
  pages={102673},
  year={2020},
  publisher={Elsevier}
}




\newpage
\appendix
\section{Experiment details}
\label{app:exp_details}

\medskip\noindent\textbf{Environment~~}The kriging baseline was implemented using SGeMS~\citep{remy2009applied}, and the tensor completion algorithm was implemented in JAX~\citep{jax2018github}. SGeMS is highly efficient, producing results for large volumes with many data points in under a minute. No Python library we tested could match SGeMS's performance, even on more powerful hardware.

\medskip\noindent\textbf{Methods' hyperparameters~~}Since our problem is formulated as a low-rank tensor approximation, the tensor completion approach has relatively few hyperparameters to tune. By contrast, significant expert effort is required to select parameters for kriging, e.g., (i) the kriging variant; (ii) the spatial correlation model (variogram type, parameters, lags); and (iii) the search ellipsoid.

We used ordinary kriging with a variogram calculated from the entire geological model. The search ellipsoid parameters were chosen as follows. First, using a $500$-well scenario, we performed a grid search over various radii and orientations to find the optimal radius and angle. We then fixed that optimal radius and, for each number of wells $\{100, 300, 500, 700\}$, performed a secondary search to fine-tune the angle.

The tensor completion algorithm was run until convergence. For each well count, we selected hyperparameters by grid search over $\rho\in\{0.1, 0.5, 0.9, 1.001, 1.01, 1.1\}$ and $\alpha \in \{10^{-3}, 10^{-2}, 10^{-1}, 1, 1.1\}$, with $\beta$ fixed at $0.1 \rho$.

\section{Reconstruction with cross-section at $z=12$}
\label{app:rec_z12}

\begin{figure*}[!h]
\normalsize
\centering
    \caption{Reconstruction results of porosity field from the SPE10 model 2. Cross-section along $z$-axis at $z = 12$ with $100$ wells. From left to right: (i) ground truth data from SPE10 model2; (ii) reconstruction with kriging; (iii) reconstruction with tensor completion; (iv) well mask.}
    \begin{center}
        \includegraphics[width=.68\textwidth]{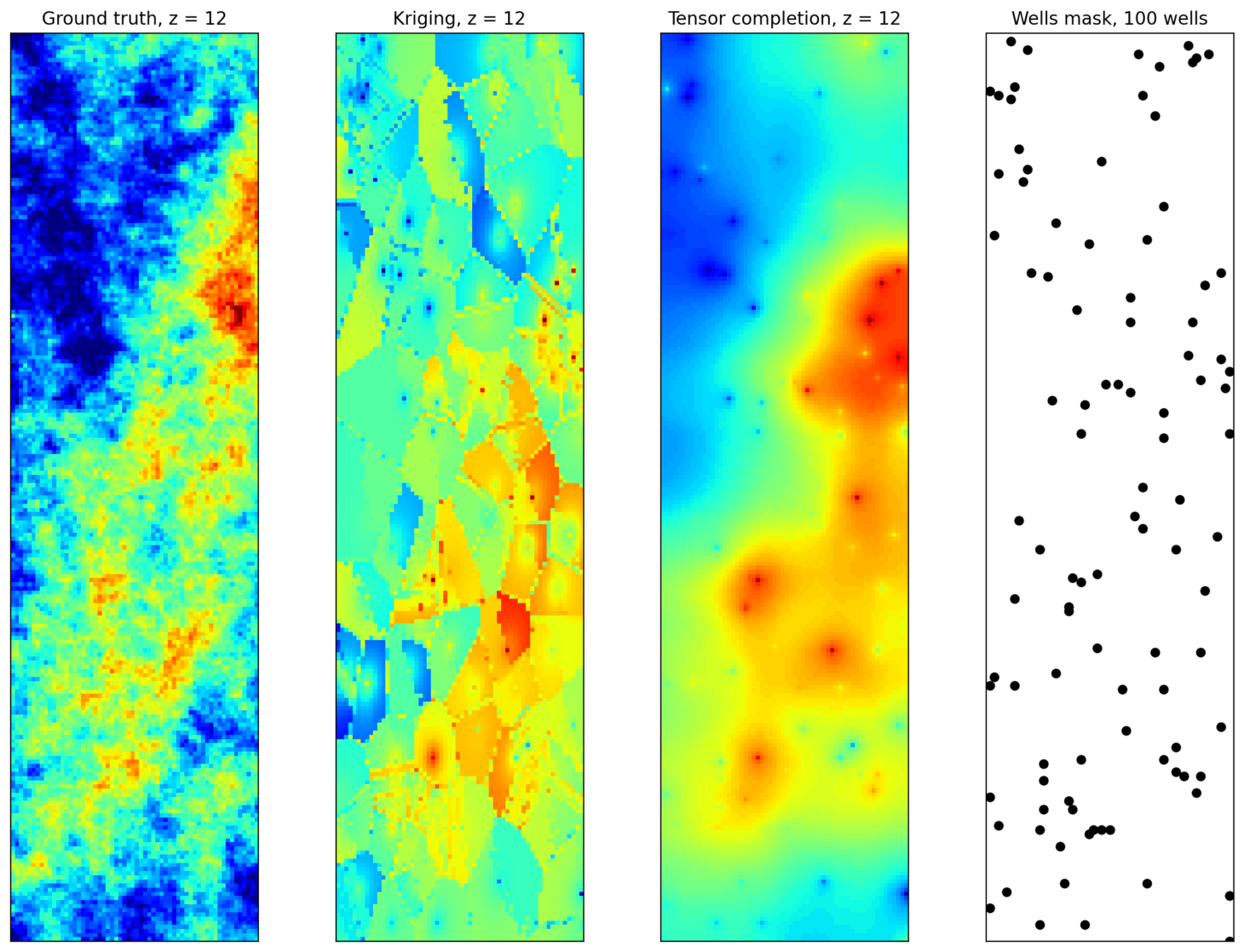}
    \end{center}
    \label{fig:app_b1}
    \vspace*{4pt}
\end{figure*}

\begin{figure*}[!h]
\normalsize
\centering
    \caption{Reconstruction results of porosity field from the SPE10 model 2. Cross-section along $z$-axis at $z = 12$ with $300$ wells. From left to right: (i) ground truth data from SPE10 model2; (ii) reconstruction with kriging; (iii) reconstruction with tensor completion; (iv) well mask.}
    \begin{center}
        \includegraphics[width=.69\textwidth]{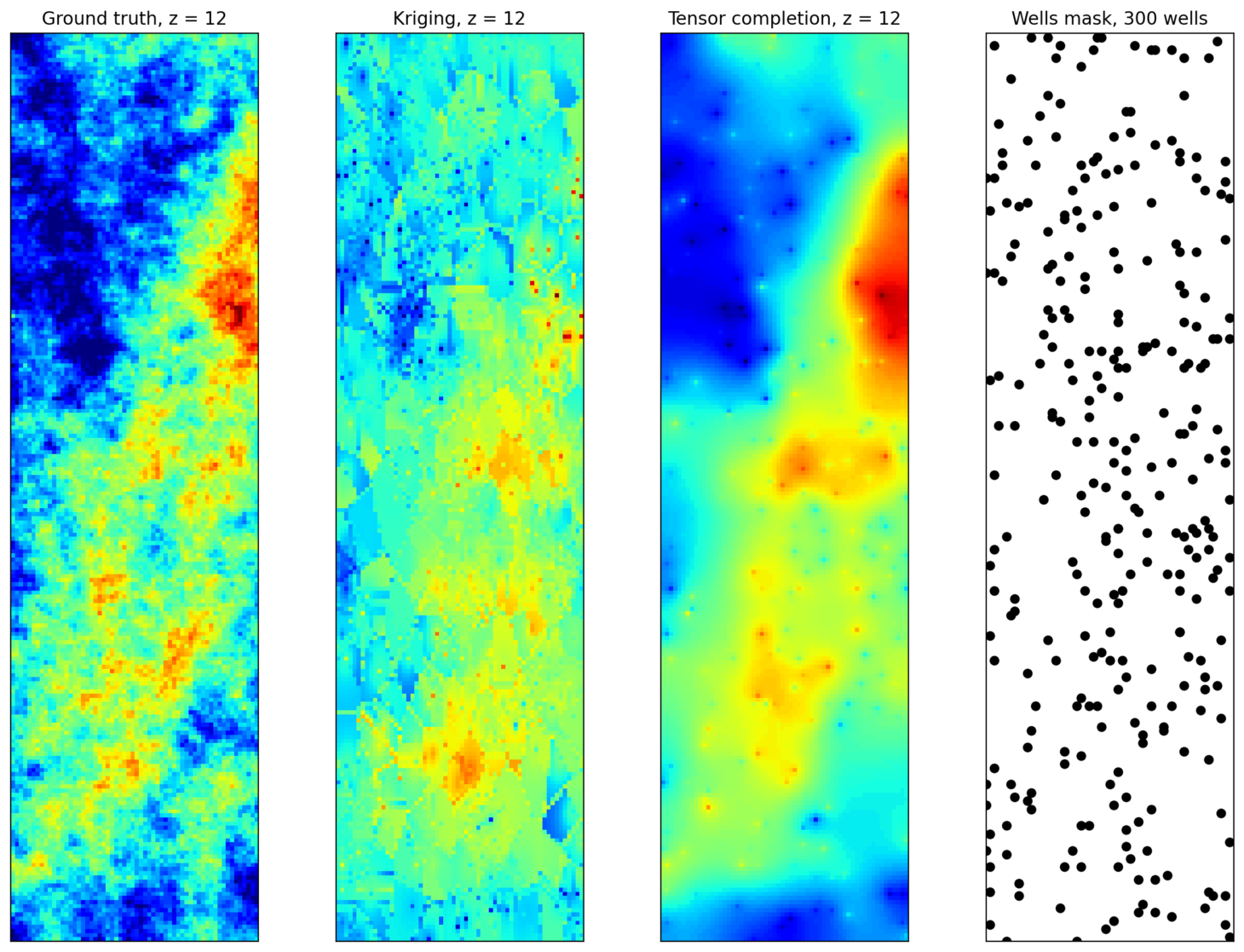}
    \end{center}
    \label{fig:app_b2}
    \vspace*{4pt}
\end{figure*}

\begin{figure*}[!h]
\normalsize
\centering
    \caption{Reconstruction results of porosity field from the SPE10 model 2. Cross-section along $z$-axis at $z = 12$ with $500$ wells. From left to right: (i) ground truth data from SPE10 model2; (ii) reconstruction with kriging; (iii) reconstruction with tensor completion; (iv) well mask.}
    \begin{center}
        \includegraphics[width=.69\textwidth]{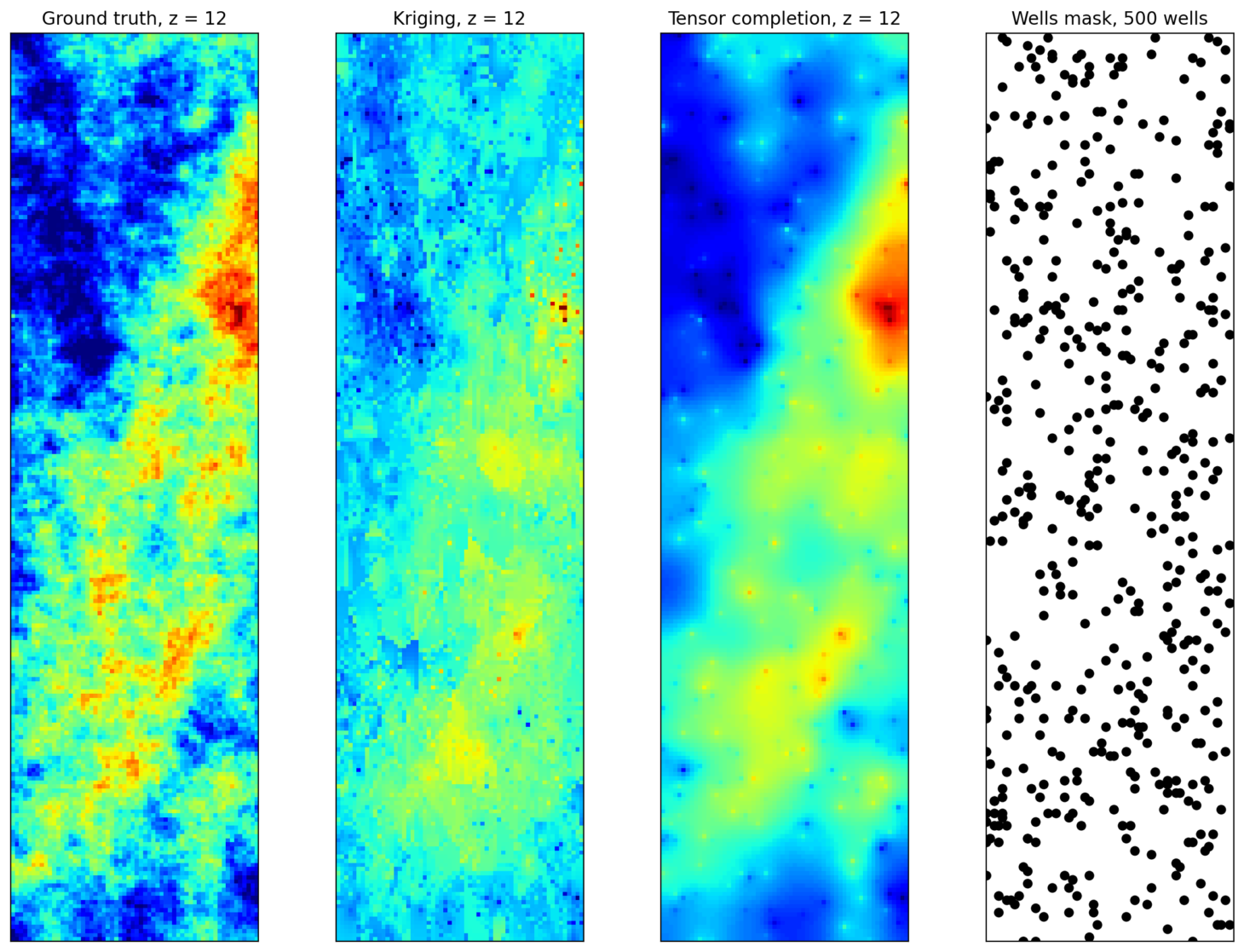}
    \end{center}
    \label{fig:app_b3}
    \vspace*{4pt}
\end{figure*}

\begin{figure*}[!h]
\normalsize
\centering
    \caption{Reconstruction results of porosity field from the SPE10 model 2. Cross-section along $z$-axis at $z = 12$ with $700$ wells. From left to right: (i) ground truth data from SPE10 model2; (ii) reconstruction with kriging; (iii) reconstruction with tensor completion; (iv) well mask.}
    \begin{center}
        \includegraphics[width=.67\textwidth]{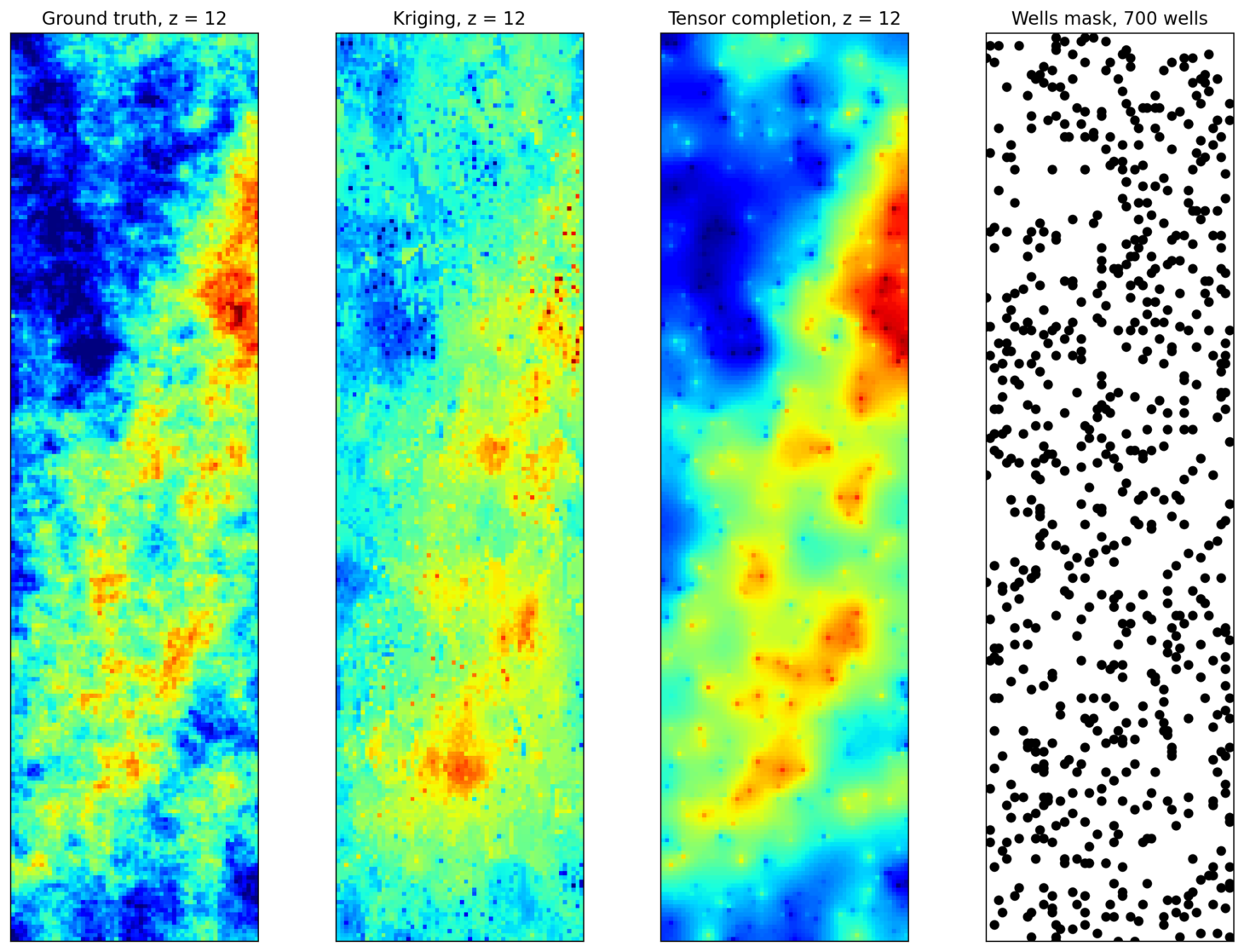}
    \end{center}
    \label{fig:app_b4}
    \vspace*{4pt}
\end{figure*}

\break
\section{Reconstruction with cross-section at $z=27$}
\label{app:rec_z27}

\begin{figure*}[!h]
\normalsize
\centering
    \caption{Reconstruction results of porosity field from the SPE10 model 2. Cross-section along $z$-axis at $z = 27$ with $100$ wells. From left to right: (i) ground truth data from SPE10 model2; (ii) reconstruction with kriging; (iii) reconstruction with tensor completion; (iv) well mask.}
    \begin{center}
        \includegraphics[width=.67\textwidth]{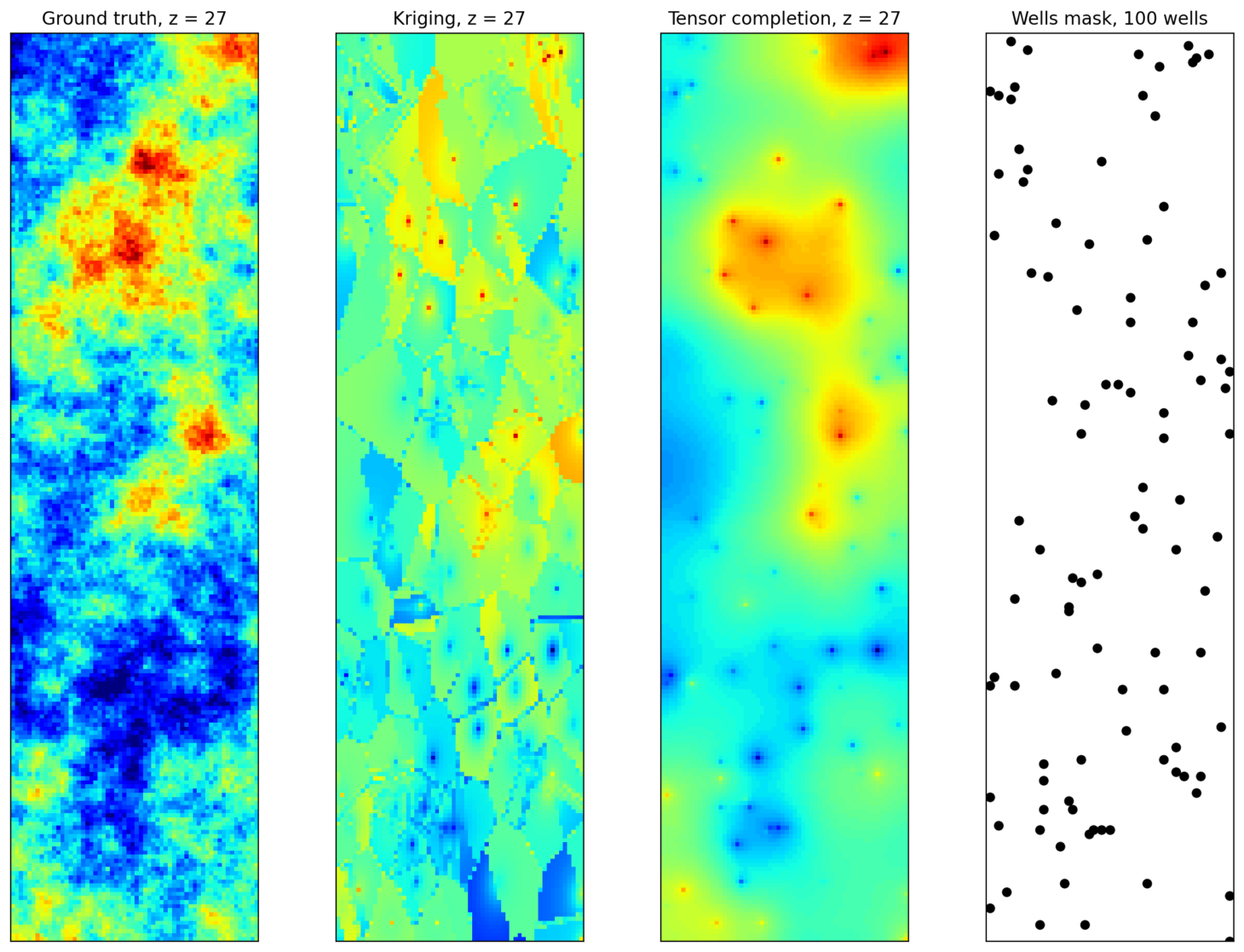}
    \end{center}
    \label{fig:app_c1}
    \vspace*{4pt}
\end{figure*}

\begin{figure*}[!h]
\normalsize
\centering
    \caption{Reconstruction results of porosity field from the SPE10 model 2. Cross-section along $z$-axis at $z = 27$ with $300$ wells. From left to right: (i) ground truth data from SPE10 model2; (ii) reconstruction with kriging; (iii) reconstruction with tensor completion; (iv) well mask.}
    \begin{center}
        \includegraphics[width=.68\textwidth]{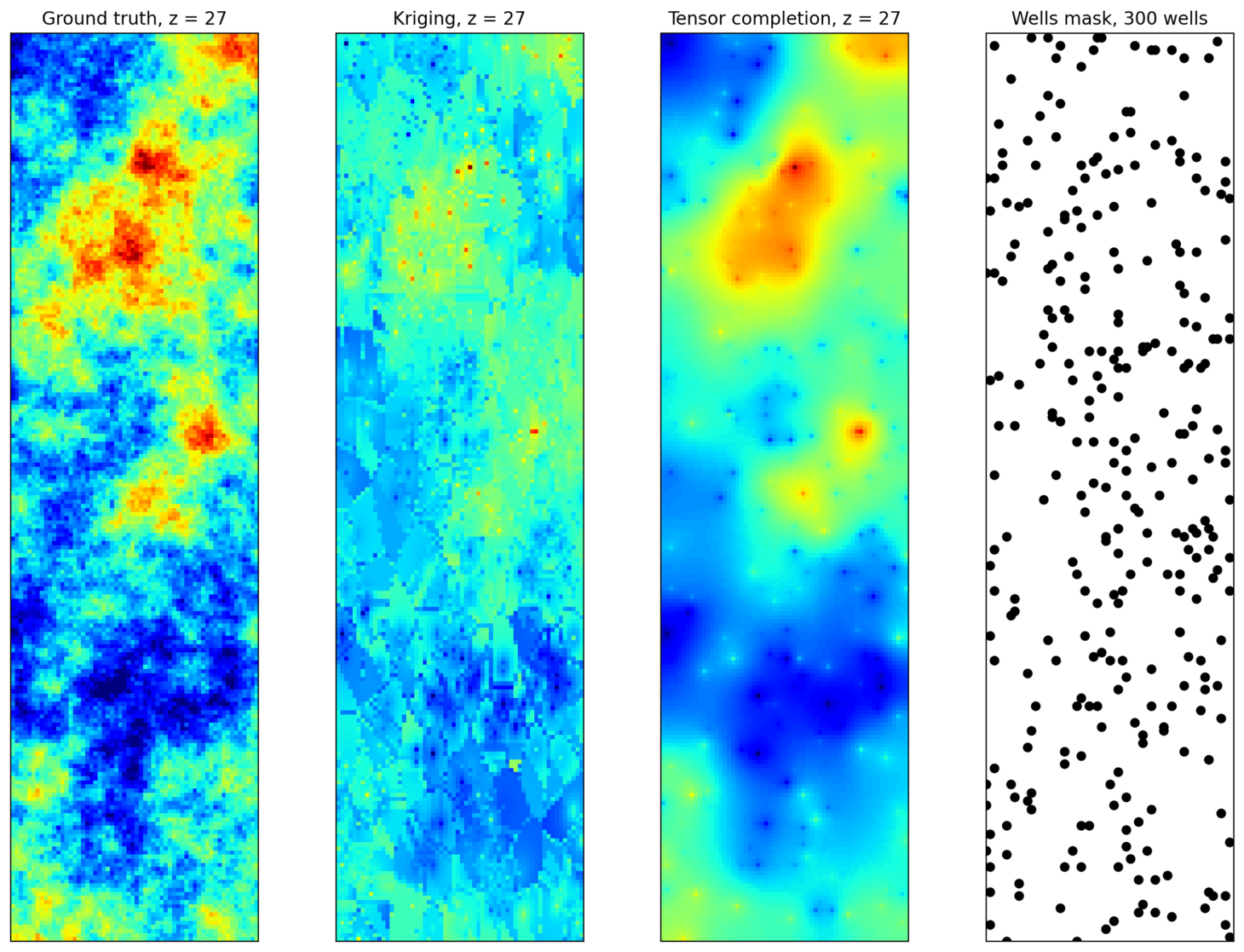}
    \end{center}
    \label{fig:app_c2}
    \vspace*{4pt}
\end{figure*}

\begin{figure*}[!h]
\normalsize
\centering
    \caption{Reconstruction results of porosity field from the SPE10 model 2. Cross-section along $z$-axis at $z = 27$ with $500$ wells. From left to right: (i) ground truth data from SPE10 model2; (ii) reconstruction with kriging; (iii) reconstruction with tensor completion; (iv) well mask.}
    \begin{center}
        \includegraphics[width=.68\textwidth]{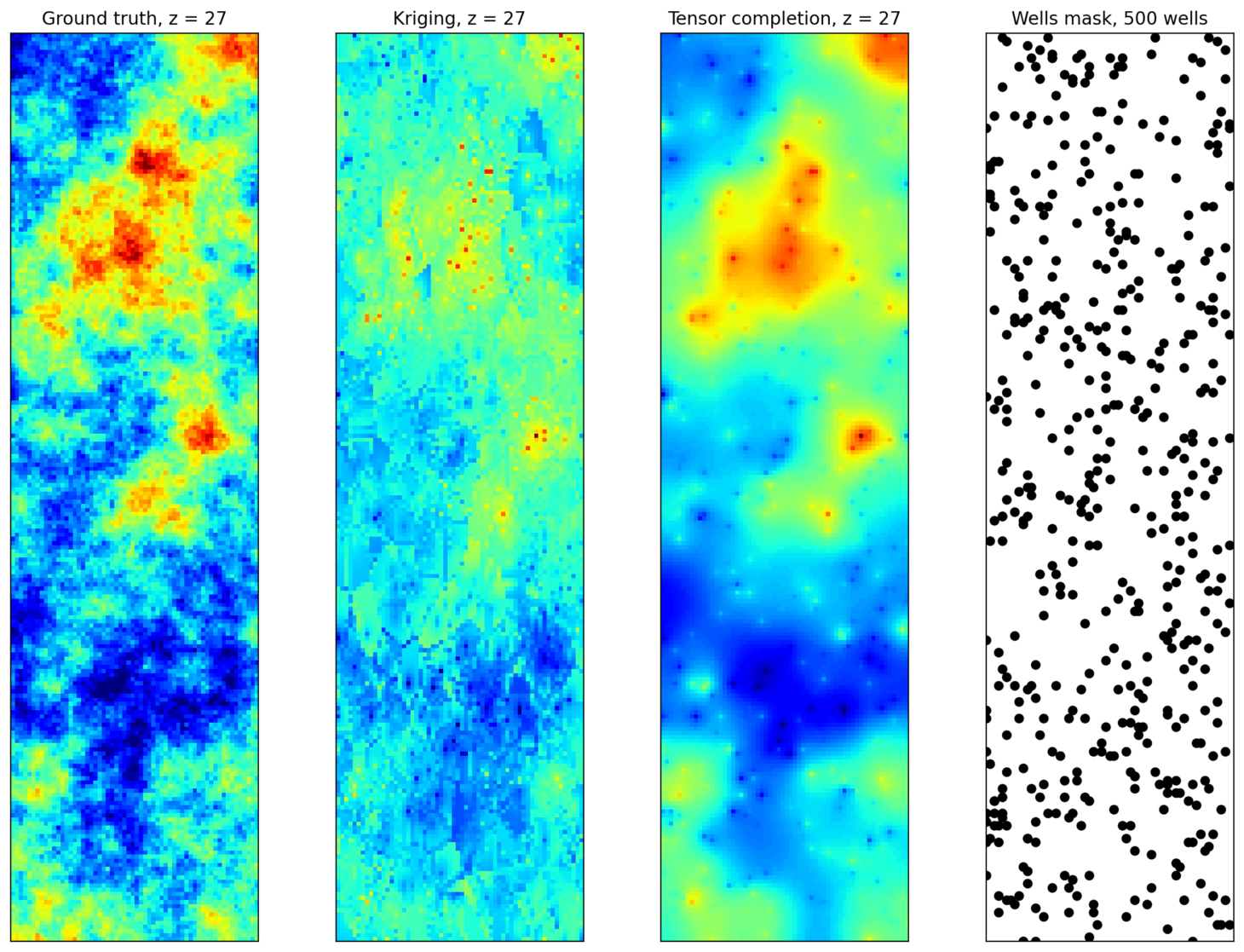}
    \end{center}
    \label{fig:app_c3}
    \vspace*{4pt}
\end{figure*}

\begin{figure*}[!h]
\normalsize
\centering
    \caption{Reconstruction results of porosity field from the SPE10 model 2. Cross-section along $z$-axis at $z = 27$ with $700$ wells. From left to right: (i) ground truth data from SPE10 model2; (ii) reconstruction with kriging; (iii) reconstruction with tensor completion; (iv) well mask.}
    \begin{center}
        \includegraphics[width=.67\textwidth]{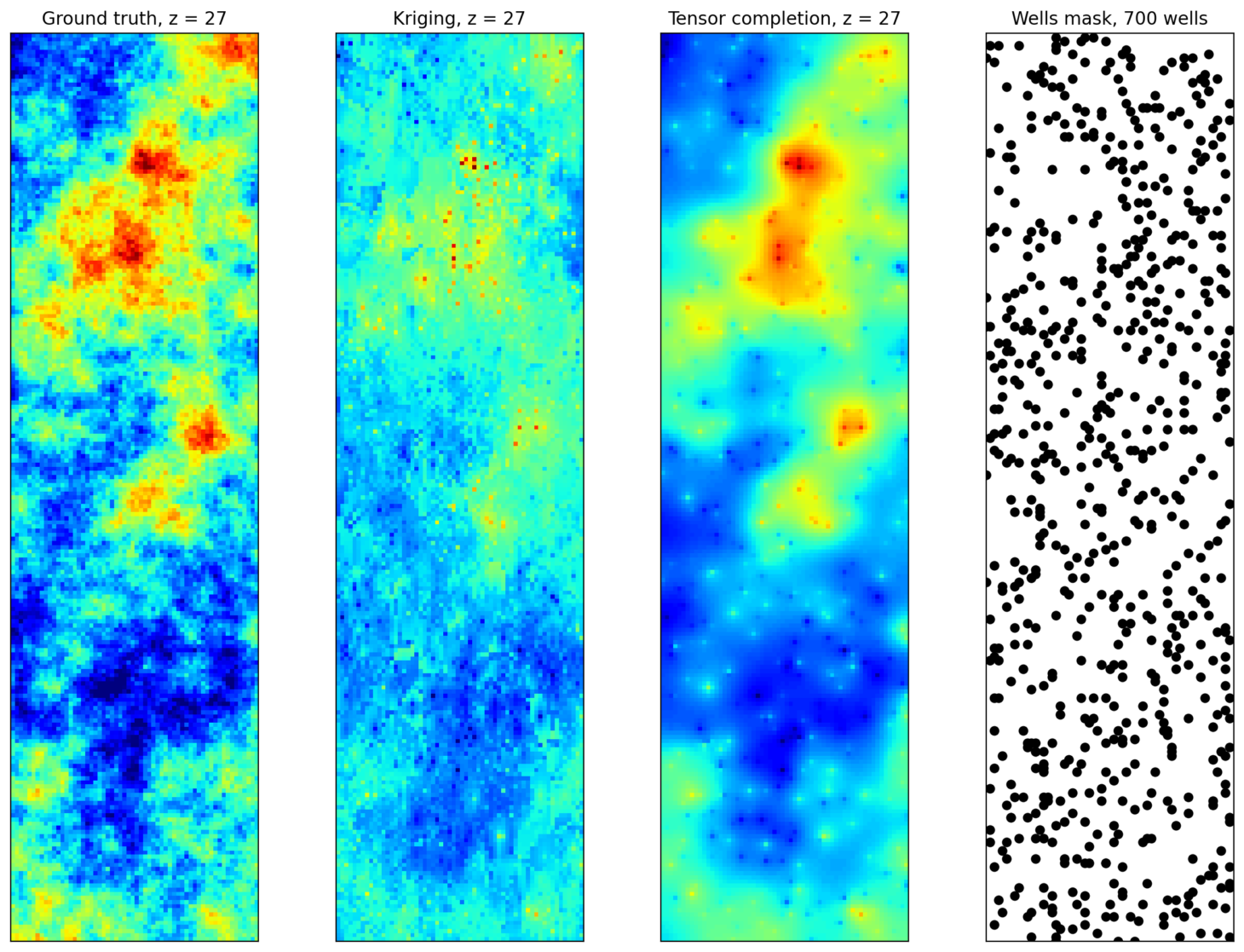}
    \end{center}
    \label{fig:app_c4}
    \vspace*{4pt}
\end{figure*}

\newpage
\section{Reconstruction with cross-section at $z=50$}
\label{app:rec_z50}

\begin{figure*}[!h]
\normalsize
\centering
    \caption{Reconstruction results of porosity field from the SPE10 model 2. Cross-section along $z$-axis at $z = 50$ with $100$ wells. From left to right: (i) ground truth data from SPE10 model2; (ii) reconstruction with kriging; (iii) reconstruction with tensor completion; (iv) well mask.}
    \begin{center}
        \includegraphics[width=.67\textwidth]{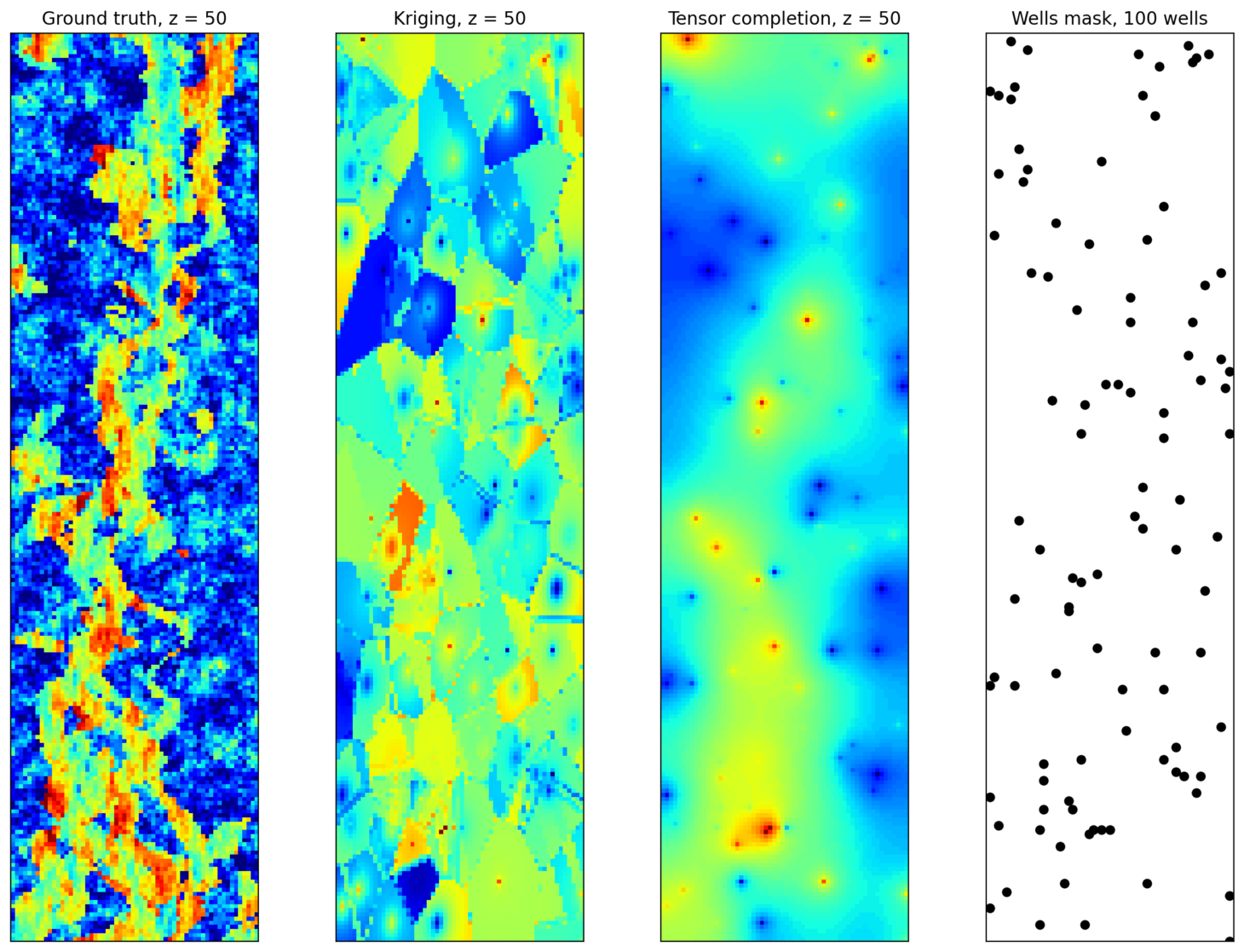}
    \end{center}
    \label{app_d1}
    \vspace*{4pt}
\end{figure*}

\begin{figure*}[!h]
\normalsize
\centering
    \caption{Reconstruction results of porosity field from the SPE10 model 2. Cross-section along $z$-axis at $z = 50$ with $300$ wells. From left to right: (i) ground truth data from SPE10 model2; (ii) reconstruction with kriging; (iii) reconstruction with tensor completion; (iv) well mask.}
    \begin{center}
        \includegraphics[width=.68\textwidth]{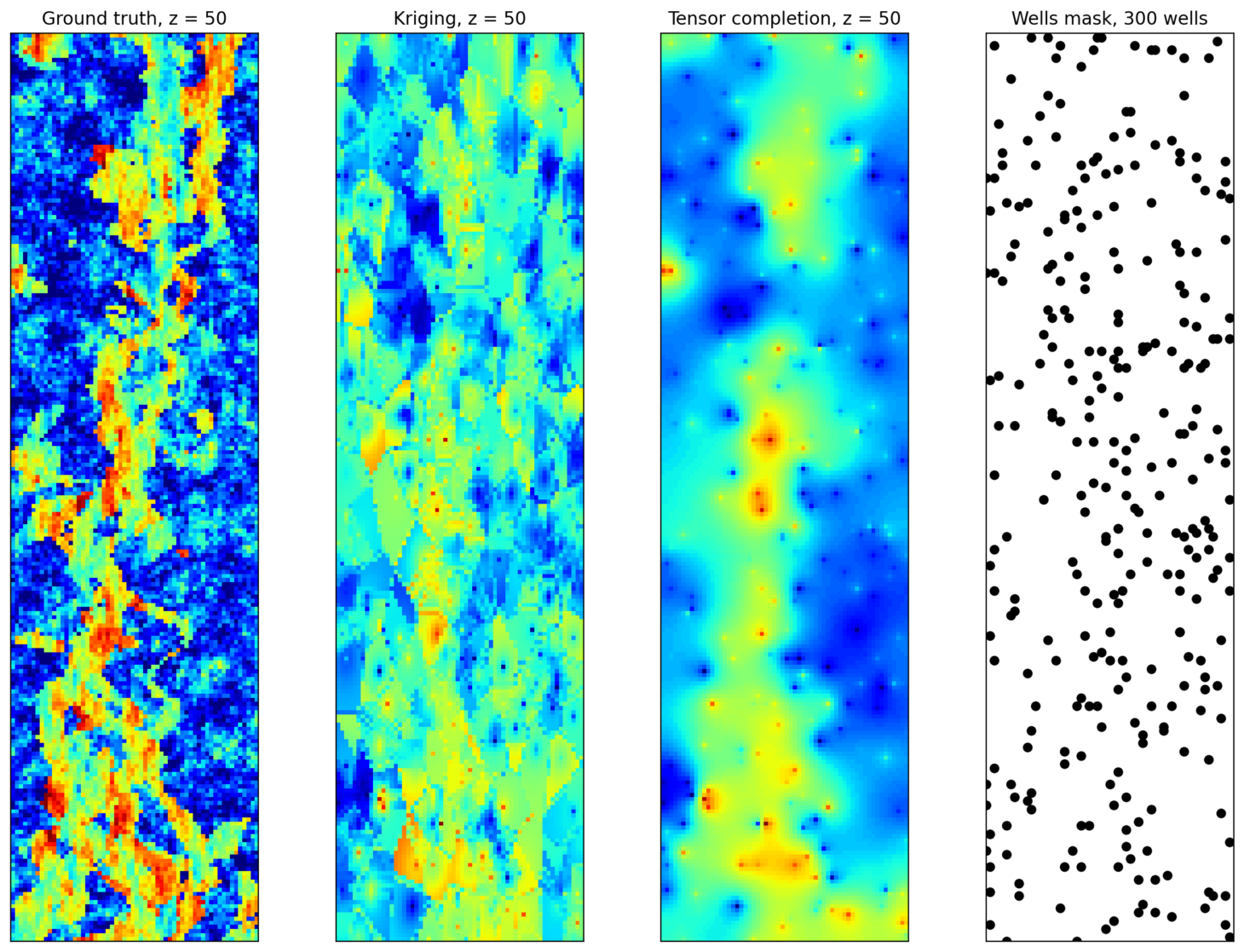}
    \end{center}
    \label{fig:app_d2}
    \vspace*{4pt}
\end{figure*}

\begin{figure*}[!h]
\normalsize
\centering
    \caption{Reconstruction results of porosity field from the SPE10 model 2. Cross-section along $z$-axis at $z = 50$ with $500$ wells. From left to right: (i) ground truth data from SPE10 model2; (ii) reconstruction with kriging; (iii) reconstruction with tensor completion; (iv) well mask.}
    \begin{center}
        \includegraphics[width=.68\textwidth]{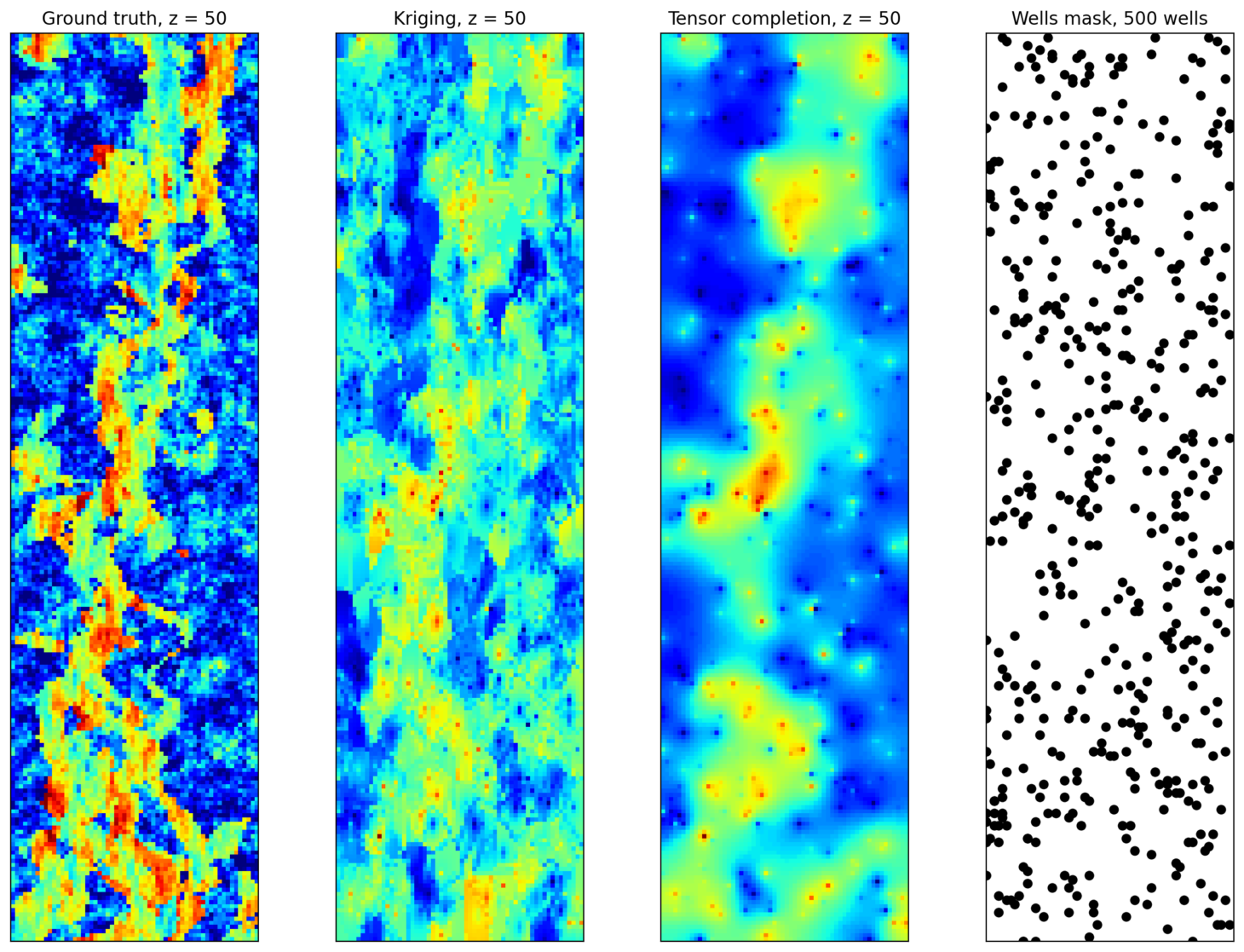}
    \end{center}
    \label{fig:app_d3}
    \vspace*{4pt}
\end{figure*}

\begin{figure*}[!h]
\normalsize
\centering
    \caption{Reconstruction results of porosity field from the SPE10 model 2. Cross-section along $z$-axis at $z = 50$ with $700$ wells. From left to right: (i) ground truth data from SPE10 model2; (ii) reconstruction with kriging; (iii) reconstruction with tensor completion; (iv) well mask.}
    \begin{center}
        \includegraphics[width=.67\textwidth]{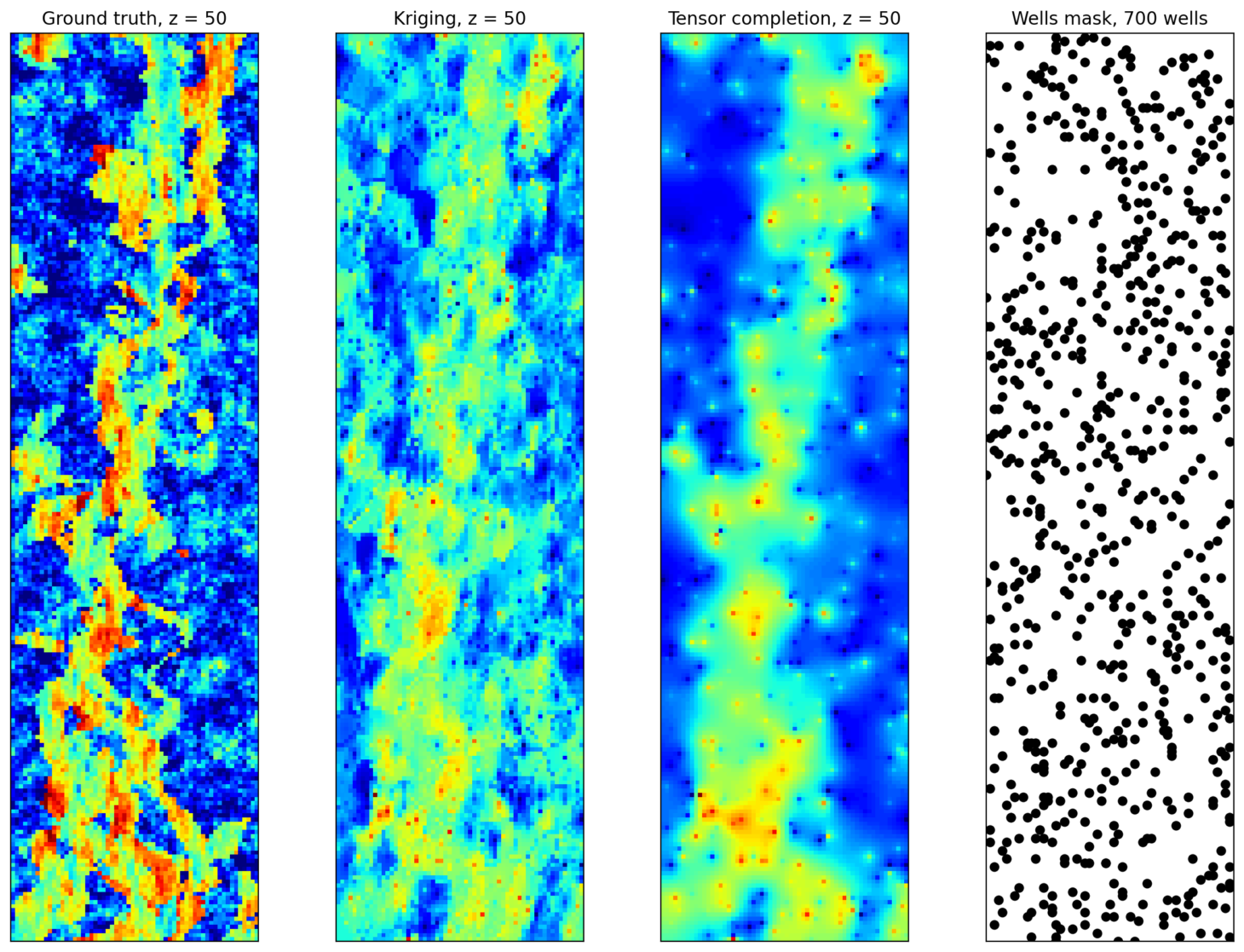}
    \end{center}
    \label{fig:app_d4}
    \vspace*{4pt}
\end{figure*}

\newpage
\section{Reconstruction with cross-section at $z=75$}
\label{app:rec_z75}

\begin{figure*}[!h]
\normalsize
\centering
    \caption{Reconstruction results of porosity field from the SPE10 model 2. Cross-section along $z$-axis at $z = 75$ with $100$ wells. From left to right: (i) ground truth data from SPE10 model2; (ii) reconstruction with kriging; (iii) reconstruction with tensor completion; (iv) well mask.}
    \begin{center}
        \includegraphics[width=.67\textwidth]{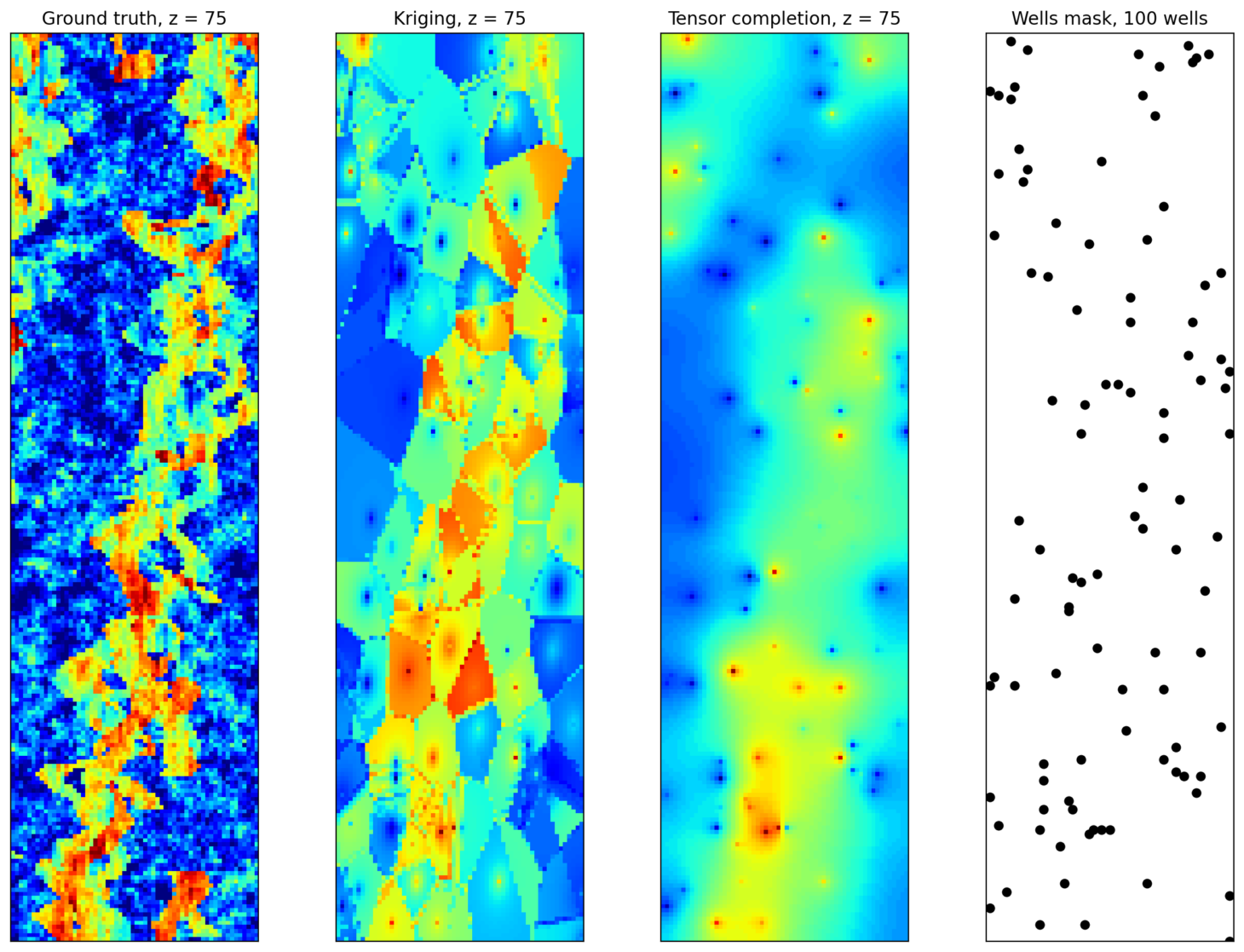}
    \end{center}
    \label{fig:app_e1}
    \vspace*{4pt}
\end{figure*}

\begin{figure*}[!h]
\normalsize
\centering
    \caption{Reconstruction results of porosity field from the SPE10 model 2. Cross-section along $z$-axis at $z = 75$ with $300$ wells. From left to right: (i) ground truth data from SPE10 model2; (ii) reconstruction with kriging; (iii) reconstruction with tensor completion; (iv) well mask.}
    \begin{center}
        \includegraphics[width=.68\textwidth]{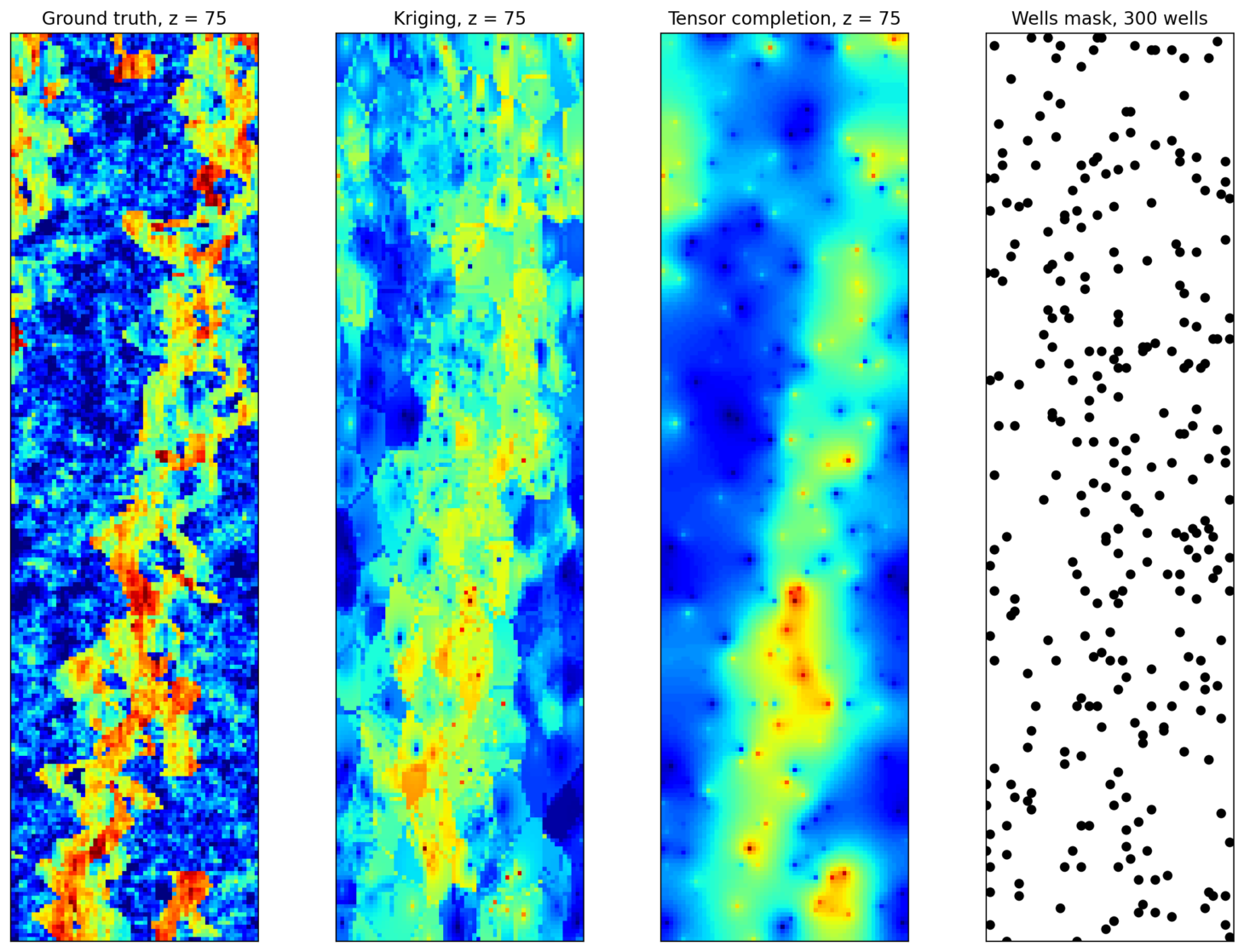}
    \end{center}
    \label{fig:app_e2}
    \vspace*{4pt}
\end{figure*}

\begin{figure*}[!h]
\normalsize
\centering
    \caption{Reconstruction results of porosity field from the SPE10 model 2. Cross-section along $z$-axis at $z = 75$ with $500$ wells. From left to right: (i) ground truth data from SPE10 model2; (ii) reconstruction with kriging; (iii) reconstruction with tensor completion; (iv) well mask.}
    \begin{center}
        \includegraphics[width=.68\textwidth]{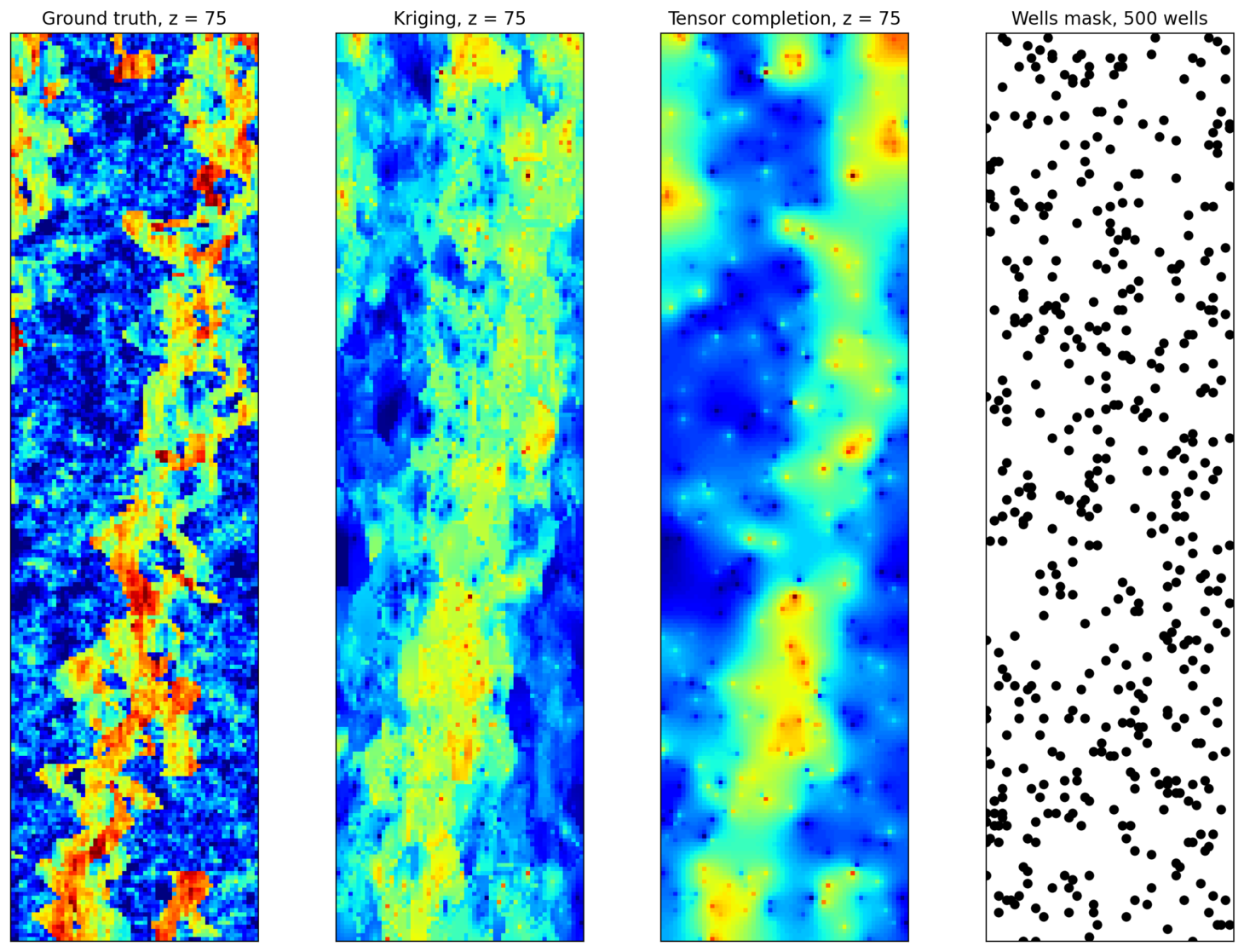}
    \end{center}
    \label{fig:app_e3}
    \vspace*{4pt}
\end{figure*}

\begin{figure*}[!h]
\normalsize
\centering
    \caption{Reconstruction results of porosity field from the SPE10 model 2. Cross-section along $z$-axis at $z = 75$ with $700$ wells. From left to right: (i) ground truth data from SPE10 model2; (ii) reconstruction with kriging; (iii) reconstruction with tensor completion; (iv) well mask.}
    \begin{center}
        \includegraphics[width=.68\textwidth]{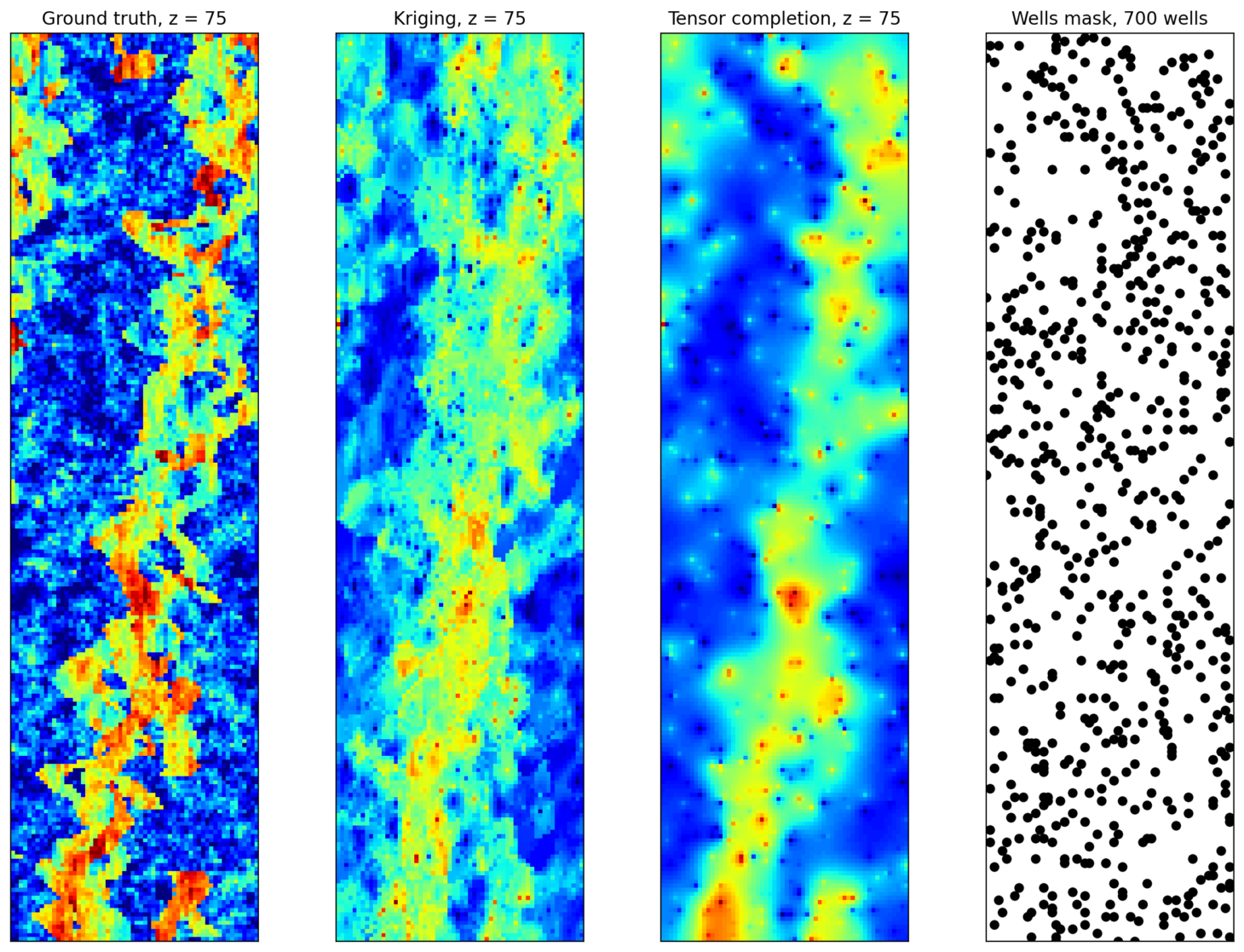}
    \end{center}
    \label{fig:app_e4}
    \vspace*{4pt}
\end{figure*}



\end{document}